%% file: main.tex
\documentclass[10pt,twocolumn,letterpaper]{article}

\usepackage[pagenumbers]{style}  

\definecolor{paperblue}{rgb}{0.21,0.49,0.74}
\usepackage[breaklinks,colorlinks,allcolors=paperblue]{hyperref}
\usepackage{multirow}

\usepackage{microtype}
\def\confName{Preprint}
\def\confYear{2027}

\title{ExplicitIR: Explicit Spatial Policy Fields for All-in-One Image Restoration}

\author{Dezhi Zhang \quad Shangrong Yang \quad Yudong Liang \quad Wenjian Wang\\[0.15em] \texttt{zhangdezhi1@sxu.edu.cn}\\[0.15em] Shanxi University}

\begin{document}

\maketitle

\begin{abstract}
\input{sections/0_abstract}
\end{abstract}

\section{Introduction}
\label{sec:intro}
\input{sections/1_introduction}

\section{Related Work}
\label{sec:related}
\begin{figure*}[t]
\centering
\includegraphics[width=\textwidth]{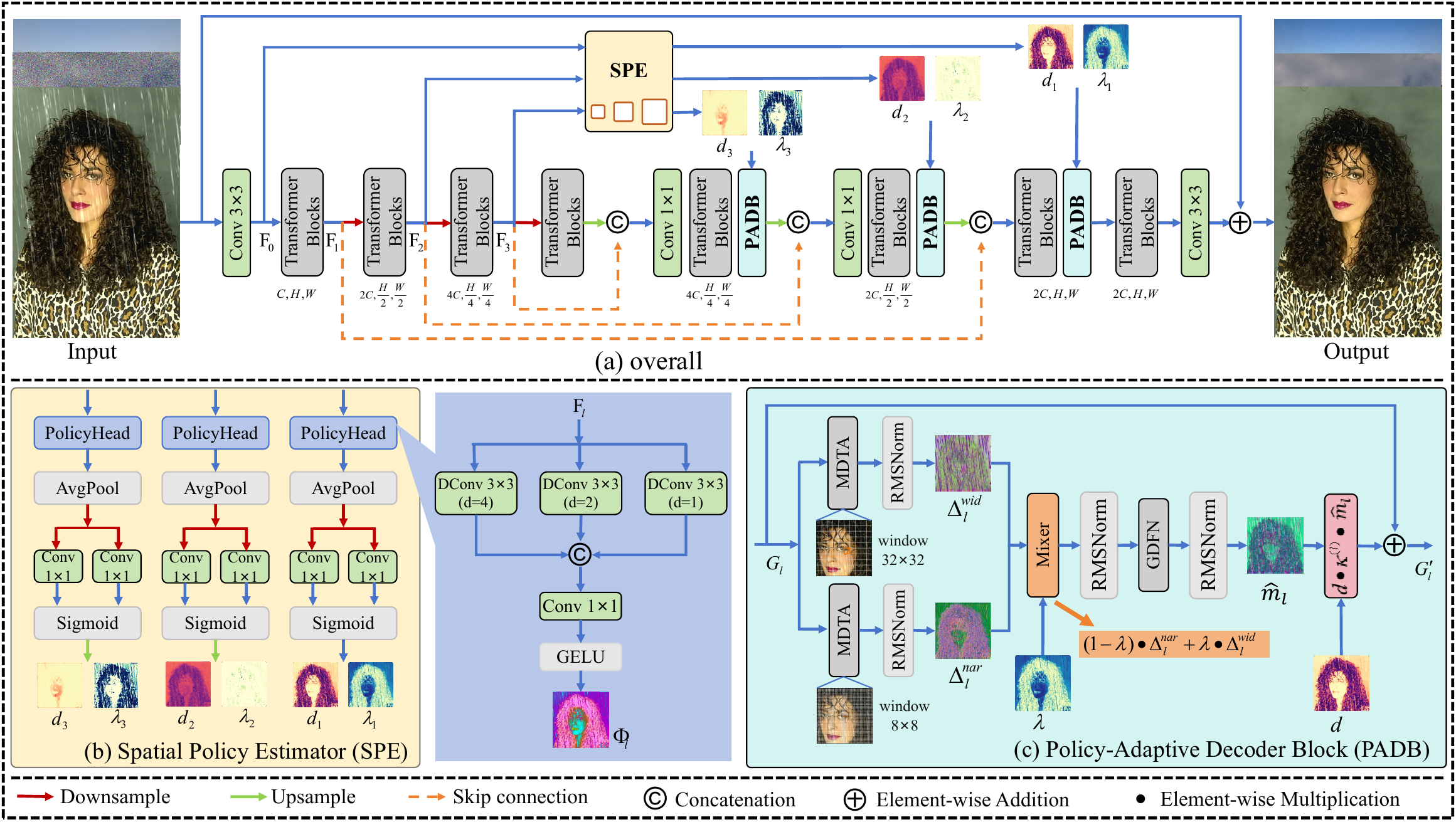}
\caption{\textbf{Overview of ExplicitIR.} \textbf{(a) Overall architecture.} A Restormer backbone equipped with a Spatial Policy Estimator (SPE) and Policy-Adaptive Decoder Blocks (PADBs). \textbf{(b) SPE module.} Multi-scale dilated convolutions (\(D_1, D_2, D_4\)) and dual independent heads predict explicit per-pixel Budget and Style fields. \textbf{(c) PADB module.} Narrow- and wide-window features are mixed via the Style field, and the result is scaled by the Budget field before injection.}
\label{fig:overview}
\end{figure*}

\input{sections/2_related_work}

\section{Method}
\label{sec:method}
\input{sections/3_method}

\section{Experiments}
\label{sec:exp}
\input{sections/4_experiments}

\section{Conclusion}
\label{sec:conc}
\input{sections/5_conclusion}




{
    \small
    \bibliographystyle{ieeenat_fullname}
    \bibliography{par_net}
}

\clearpage
\appendix
\input{appendix}

\end{document}

%% file: sections/0_abstract.tex
All-in-one image restoration aims to recover clean images blindly from inputs with various degradation types and levels using a single model.
However, the adaptation policy in most existing methods remains implicitly encoded in feature representations, leaving the restoration decision neither explicitly represented nor independently controllable.
Since real-world degradations vary spatially, applying uniform updates across the entire image leads to over-smoothing of clean areas and under-correction of localized degradations.
To address these limitations, we present ExplicitIR, which represents the restoration policy as two explicit per-pixel fields: Budget, which determines where and how strongly to update, and Style, which selects between narrow and wide context windows.
Specifically, ExplicitIR incorporates a Spatial Policy Estimator (SPE) to predict these fields, explicitly representing the restoration policy and capturing spatially-varying degradations, which are then applied by Policy-Adaptive Decoder Blocks (PADB) through adaptive fusion of narrow- and wide-window responses, enabling independent control and precise local modulation.
This design enables spatially-adaptive restoration in a single forward pass, without requiring task labels, prompts, or test-time degradation estimators.
Extensive experiments on three-task and five-task benchmarks demonstrate that ExplicitIR achieves performance competitive with the strongest all-in-one baselines.

%% file: sections/1_introduction.tex
Image restoration aims to recover a high-quality image from a degraded observation. With large-scale training data, deep networks have learned strong priors, and single-degradation restoration---covering noise, rain, and haze---has reached a high standard~\cite{zamir2021multi,liang2021swinir,chen2022nafnet,zamir2022restormer,wang2022uformer,sun2024histoformer}. However, real-world images are rarely degraded by a single factor; instead, multiple degradations such as noise, blur, and adverse weather often coexist~\cite{guo2024onerestore,zamfir2025moceir,he2026dacgir}. Deploying a separate model for each degradation type is impractical, especially on limited hardware. Such a system requires inferring the degradation type, routing to the correct model, and enduring memory and latency overheads~\cite{li2022airnet,potlapalli2023promptir,conde2024instructir}. All-in-one image restoration (AiOIR) aims to recover clear images from various degradation types and levels with a unified model~\cite{li2022airnet,potlapalli2023promptir,tian2025dfpir}.

\begin{figure*}[t]
\centering
\includegraphics[width=\textwidth]{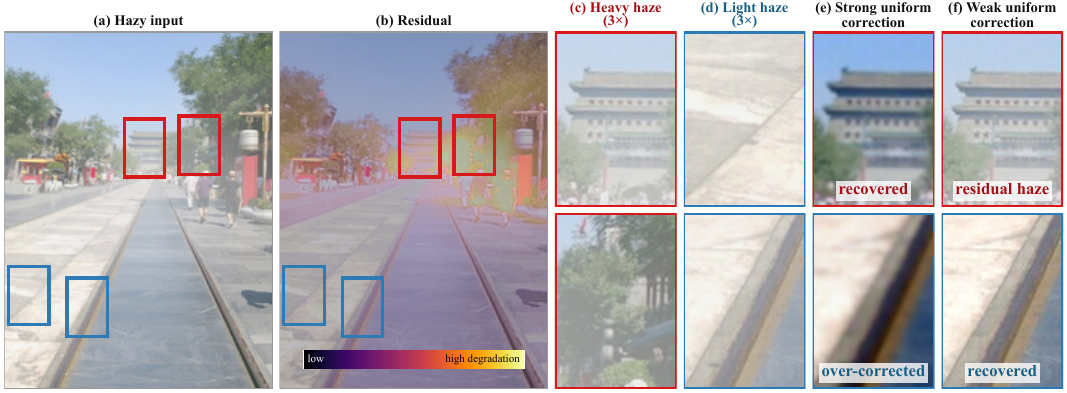}
\caption{Motivation. (a)~A hazy street scene. Several regions are highlighted with bounding boxes: the red box marks a heavily degraded region, while the blue box marks a mildly degraded region. (b)~The absolute residual map against the clean reference, where brighter areas indicate more severe degradation. (c,~d)~Magnified views of the boxed regions. (e)~One strong uniform correction applied everywhere: the tower is recovered, but the pavement is over-corrected. (f)~One weak uniform correction: the pavement is preserved, but haze remains over the tower. This demonstrates that a single uniform strength cannot serve both regions, and the correction must vary with location.}
\label{fig:teaser}
\end{figure*}

Existing AiOIR methods can be broadly grouped into three categories. Prompt-based methods condition a shared backbone on a global degradation representation~\cite{li2022airnet,zhang2023idr,potlapalli2023promptir,conde2024instructir}, which is applied uniformly across all locations, failing to accommodate spatially varying degradations. Architecture-oriented methods instead modify the network computation or parameter space~\cite{tian2025dfpir,zamfir2025moceir,du2026symunet}, but their inherent image-level adaptations lack the spatial resolution to model non-uniform corruption. Spatially selective methods introduce location-specific operations~\cite{cui2025adair,he2026dacgir,he2026cea}. However, their spatial masks remain tightly coupled with the feature computation itself, preventing independent control over the update strength at each location. Consequently, they over-smooth clean areas and under-correct corrupted regions.

As illustrated in Fig.~\ref{fig:teaser}, restoration demand varies spatially within a single image, as different regions often exhibit distinct degradation types and severity levels. To handle such spatial heterogeneity, an effective restoration model should (i) decide where and how strongly to update each location, (ii) select between local and global context per location, and (iii) explicitly represent these two decisions. Crucially, because update strength and context are highly coupled, relying on a single global configuration inevitably leads to a coarse trade-off. Explicitly decoupling and representing them is therefore essential.

To address the aforementioned issues, we propose ExplicitIR, an all-in-one restoration framework that represents the restoration policy as two explicit per-pixel fields: Budget, which determines where and how strongly to update each location, and Style, which selects between narrow and wide context windows. Specifically, a Spatial Policy Estimator (SPE) extracts these fields directly from Restormer~\cite{zamir2022restormer} encoder features, allowing the network to accurately localize corrupted regions without perturbing the underlying feature representations. Policy-Adaptive Decoder Blocks (PADB) then execute this policy, guaranteeing that the update magnitude strictly respects the predicted Budget and effectively preventing over-correction in clean areas and under-correction in damaged ones. This design achieves fine-grained spatial control while maintaining high efficiency, allowing the model to operate in a single forward pass at inference without any task labels, prompts, or degradation estimators. The main contributions of this work are as follows:

\begin{itemize}
\item We propose ExplicitIR, an all-in-one restoration framework that explicitly represents the restoration policy as two per-pixel fields: Budget and Style. This explicit representation allows the adaptation decision to be directly inspected and edited.
\item We design a Spatial Policy Estimator (SPE) to predict both fields from multi-scale encoder features, and Policy-Adaptive Decoder Blocks (PADB) to execute them. This decoupled architecture ensures that update strength and context are independently controlled at each location, enabling highly precise localized restoration and preserving the overall image fidelity.
\item Extensive experiments on the three-task and five-task benchmarks demonstrate that ExplicitIR achieves performance competitive with the strongest all-in-one baselines.
\end{itemize}

%% file: sections/2_related_work.tex
\paragraph{Prompt-based all-in-one restoration.}
Early approaches extract a degradation representation to modulate the network, such as contrastive learning in AirNet~\cite{li2022airnet}, ingredient decomposition in IDR~\cite{zhang2023idr}, and composite-degradation descriptors in OneRestore~\cite{guo2024onerestore}. Subsequent methods adopt explicit cues, including visual prompts in PromptIR~\cite{potlapalli2023promptir} and Wu et al.~\cite{wu2026cpl}, text instructions in InstructIR~\cite{conde2024instructir}, weight modulation in InstructIPT~\cite{tian2024instructipt}, diffusion steering in DeFusion~\cite{luo2025defusion} and FAPEIR~\cite{liu2026fapeir}, and multimodal LLM embeddings in MMFE-IR~\cite{lee2026mmfeir}. Related efforts also explore cyclic prompting~\cite{liao2025cyclicprompt} and degradation perception enhancement~\cite{zhang2026perceiveir}. Despite their diversity, these methods compute a compact global representation once per image, which bottlenecks localized degradation evidence and precludes fine-grained spatial adaptation~\cite{he2026cea}.

\paragraph{Architecture-based all-in-one restoration.}
Architecture-based methods instead modify the network computation or parameter space. Representative examples include feature perturbation in DFPIR~\cite{tian2025dfpir}, expert routing in MoCE-IR~\cite{zamfir2025moceir}, and low-rank degradation embeddings in DaAIR~\cite{zamfir2024daair}. UP-Restorer~\cite{liu2025uprestorer} unrolls the restoration process, while SymUNet~\cite{du2026symunet} adopts a symmetric backbone, FoundIR~\cite{li2025foundir} and FoundIR-v2~\cite{chen2025foundirv2} scale training data and optimize pre-training mixtures, and DVANet~\cite{tu2026dvanet} casts unified restoration as deep unfolding. However, because these mechanisms still rely on global feature representations, they remain image-level adaptations, preventing them from resolving spatially non-uniform corruption at the pixel level~\cite{he2026cea}.

\paragraph{Spatially selective restoration.}
Spatially selective methods introduce location-specific operations to handle spatial non-uniformity, from patch routing in Path-Restore~\cite{yu2022pathrestore} and semantic priors in SFT-GAN~\cite{wang2018sftgan}, to frequency-adaptive modulation in AdaIR~\cite{cui2025adair}. Within AiOIR, DACG-IR~\cite{he2026dacgir} gates features with degradation-aware masks, and CEA~\cite{he2026cea} assembles instance-conditioned low-rank residuals. While providing spatial awareness, these masks remain embedded within the feature computation, leaving the restoration policy implicit in the network weights. As a result, the update strength is not directly measurable or supervisable. In contrast, ExplicitIR externalizes the policy into two explicit per-pixel fields with a separate executor, making the adaptation directly inspectable and controllable.

%% file: sections/3_method.tex
ExplicitIR represents the adaptation policy as two per-pixel fields: Budget \(d\) sets the update strength at each location, and Style \(\lambda\) selects between narrow and wide context.
Fig.~\ref{fig:overview} shows the architecture: a Restormer~\cite{zamir2022restormer} encoder-decoder, a Spatial Policy Estimator (SPE) that predicts both fields from encoder features, and Policy-Adaptive Decoder Blocks (PADB) that apply them at each decoder stage.
Sec.~\ref{sec:spe} describes SPE, Sec.~\ref{sec:padb} describes PADB, and Sec.~\ref{sec:training} describes the training objective.

\subsection{Overall Pipeline}
\label{sec:overview}
ExplicitIR extracts shallow features with a convolution and uses a four-level Restormer encoder-decoder~\cite{zamir2022restormer}.
SPE reads three of these features, \(F_0\), \(F_2\) and \(F_3\): the stem output and the second and third encoder levels.
From these features, SPE predicts stage-wise fields \(\{d_l,\lambda_l\}_{l=1}^{3}\) (Sec.~\ref{sec:spe}).
A PADB appended to each of the three shallowest decoder stages applies the corresponding fields to the stage output (Sec.~\ref{sec:padb}).
The final refinement stage and the residual head \(\hat{I}=I+R\) are unchanged, so ExplicitIR inherits the backbone capacity and adds only the policy path.
Figure~\ref{fig:overview} summarizes the pipeline.

\subsection{Spatial Policy Estimator (SPE)}
\label{sec:spe}
Haze strength follows scene depth, and rain or noise follows local structure.
A single receptive field would tie the estimate to one scale, so the fields must be estimated from multiple scales and depths.

Each branch reads one encoder feature, \(F_0\), \(F_2\) or \(F_3\), and captures its scale with a bank of dilated depthwise convolutions:
\begin{equation}
\label{eq:phi}
\phi_l=\varphi\big(W_{\mathrm{fuse}}\,[\,D_{1}(F_l)\,\|\,D_{2}(F_l)\,\|\,D_{4}(F_l)\,]\big),
\end{equation}
where \(D_r\) denotes a depthwise \(3{\times}3\) convolution with dilation rate \(r\), \([\cdot\|\cdot\|\cdot]\) is channel concatenation, \(W_{\mathrm{fuse}}\) is a \(1{\times}1\) convolution, and \(\varphi\) is the GELU non-linearity.
Two independent \(1{\times}1\) heads then predict
\begin{equation}
\label{eq:spe}
\begin{gathered}
d_l=\sigma\!\big(H_d^{(l)}(\phi_l)\big),\\
\lambda_l=\sigma\!\big(H_\lambda^{(l)}(\phi_l)\big),
\end{gathered}
\end{equation}
where \(\sigma\) denotes the sigmoid.
The three fields are estimated at resolutions \(H/2\), \(H/4\) and \(H/4\), and resampled to \(H\), \(H/2\) and \(H/4\) before they act on the matching decoder stage.

\begin{figure}[t]
\centering
\includegraphics[width=\columnwidth]{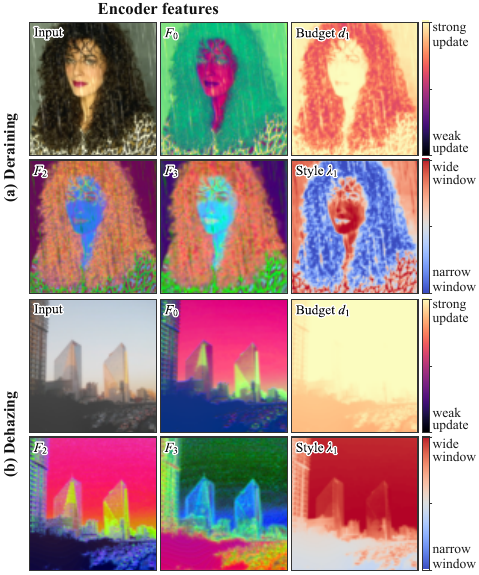}
\caption{\textbf{Mechanism visualization of the Spatial Policy Estimator (SPE).} Given the encoder features \(F_0\), \(F_2\) and \(F_3\), the Budget field \(d\) controls the update strength, and the Style field \(\lambda\) determines the window size. \textbf{(a) Deraining:} \(d\) concentrates on localized rain streaks. \textbf{(b) Dehazing:} \(d\) distributes broadly over the homogeneous haze.}
\label{fig:spe}
\end{figure}

SPE predicts both fields from a shared multi-scale representation, so Budget and Style are estimated jointly rather than through separate pathways.
As visualized in Fig.~\ref{fig:spe}, the predicted Budget field tightly concentrates on localized rain streaks, but distributes globally over the homogeneous haze, demonstrating its ability to capture spatially-varying degradation patterns.

\subsection{Policy-Adaptive Decoder Block (PADB)}
\label{sec:padb}
Style needs alternatives to select from: a single window width would fix the context of every update at one scale.
Direct mixing would also let the update amplitude follow the attention response, so the fields would no longer control the update strength.

Each block runs the Restormer attention operator, Multi-Dconv Head Transposed Attention (MDTA)~\cite{zamir2022restormer}, in a narrow-window and a wide-window form.
Both forms take each location's channel statistics over a non-overlapping neighborhood of side \(w_{\mathrm{nar}}\) and \(w_{\mathrm{wid}}\) instead of the whole feature map, and are otherwise identical to the full-map operator (Appendix~A).
Both branches are normalized to unit per-location RMS, mixed under Style, and renormalized after synthesis:
\begin{equation}
\label{eq:delta}
\begin{gathered}
\Delta_l^{\mathrm{nar}}=\frac{\mathrm{Narrow}(G_l)}{\mathrm{RMS}(\mathrm{Narrow}(G_l))},\\
\Delta_l^{\mathrm{wid}}=\frac{\mathrm{Wide}(G_l)}{\mathrm{RMS}(\mathrm{Wide}(G_l))},\\
m_l=\mathrm{norm}\!\big((1-\lambda_l)\Delta_l^{\mathrm{nar}}+\lambda_l\Delta_l^{\mathrm{wid}}\big),\\
\hat{m}_l=\mathrm{norm}\!\big(\mathrm{GDFN}(m_l)\big).
\end{gathered}
\end{equation}
where \(\mathrm{norm}(x)=x/\mathrm{RMS}(x)\), \(\mathrm{RMS}(\cdot)\) denotes the per-location root mean square over channels computed with a stabilizer \(\epsilon\) (Appendix~A), \(\mathrm{Narrow}(\cdot)\) and \(\mathrm{Wide}(\cdot)\) are the two windowed forms of the same operator, \(\mathrm{GDFN}\) is the Gated-Dconv Feed-Forward Network of the backbone, and \(G_l\) is the decoder feature entering the block.
The first two normalizations ensure \(\|\Delta_l^{\mathrm{nar}}\|_{\mathrm{RMS}}=\|\Delta_l^{\mathrm{wid}}\|_{\mathrm{RMS}}=1\), which makes \(\lambda_l\) a true convex mix and keeps the two branches on the same scale.
The third is applied after the GDFN, so \(\|\hat{m}_l\|_{\mathrm{RMS}}=1\) regardless of the synthesis output.
The block output is
\begin{equation}
\label{eq:padb}
G_l'=G_l+d_l\,\kappa^{(l)}\,\hat{m}_l,
\end{equation}
where \(\kappa^{(l)}\) is a per-stage constant that fixes the update amplitude.
Eq.~\eqref{eq:padb} scales a direction that is already normalized to unit RMS, so the update amplitude is determined solely by \(d_l\,\kappa^{(l)}\).
If Style also carried amplitude, \(\|G_l'-G_l\|_{\mathrm{RMS}}\) would depend on \(\lambda_l\), and the bound in Eq.~\eqref{eq:bound} would not hold.
Consequently,
\begin{equation}
\label{eq:bound}
\begin{gathered}
\|G_l'-G_l\|_{\mathrm{RMS}}=d_l\,\kappa^{(l)},\\
d_l=0\Rightarrow G_l'=G_l .
\end{gathered}
\end{equation}

\begin{figure}[t]
\centering
\includegraphics[width=\columnwidth]{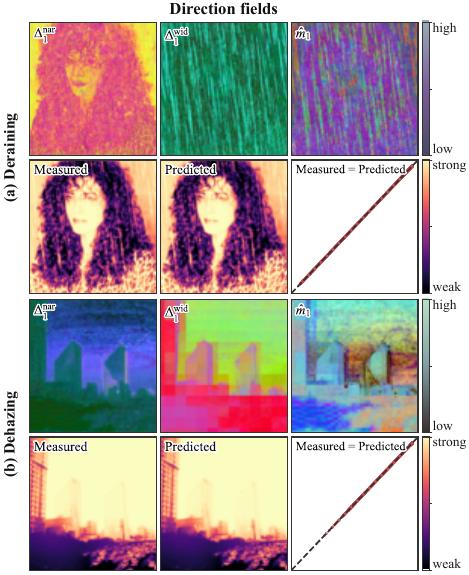}
\caption{\textbf{Mechanism verification of the Policy-Adaptive Decoder Block (PADB) on (a) deraining and (b) dehazing inputs.} \textbf{Top row:} Direction fields of the narrow-window branch \(\Delta_1^{\mathrm{nar}}\), the wide-window branch \(\Delta_1^{\mathrm{wid}}\), and their Style-mixed output \(\hat{m}_1\) under a shared PCA basis. \textbf{Bottom row:} The measured update magnitude, the predicted magnitude \(d_1\kappa^{(1)}\), and their scatter plot. The points align strictly on the \(y=x\) diagonal.}
\label{fig:padb}
\end{figure}

By Eq.~\eqref{eq:bound}, the update never exceeds \(\kappa^{(l)}\) in RMS at any location, so Budget alone sets its amplitude, and a location with vanishing Budget passes through the decoder unchanged.
Fig.~\ref{fig:padb} provides a visual verification of this bound, where the measured update magnitudes align strictly along the \(y=x\) diagonal with the predictions \(d_l \kappa^{(l)}\), confirming that the injected update is governed exclusively by the Budget field.

\subsection{Training Objective}
\label{sec:training}
Every model reported in this paper is trained with a single reconstruction loss,
\begin{equation}
\label{eq:total}
\mathcal{L}=\|\hat{I}-I_{\mathrm{gt}}\|_1,
\end{equation}
applied in the RGB domain.
No auxiliary consistency loss or external guidance is used, allowing the network to learn the policy fields directly from the restoration signal.

\subsection{Analysis}
\label{sec:analysis}
Eq.~\eqref{eq:bound} implies a ratio invariance.
For any two locations \(i\) and \(j\) at the same decoder stage with non-vanishing Budget,
\begin{equation}
\label{eq:ratio}
\frac{\|G_l'(i)-G_l(i)\|_{\mathrm{RMS}}}{\|G_l'(j)-G_l(j)\|_{\mathrm{RMS}}}=\frac{d_l(i)}{d_l(j)}.
\end{equation}
This ratio is independent of the backbone's residual scale and of the training dynamics, so the output displacement is strictly proportional to the Budget at each location.
Consequently, the two fields govern orthogonal aspects of the update: Budget controls the magnitude, while Style selects the direction. Appendix~B visualizes how explicitly manipulating these fields at inference confirms this separation.

%% file: sections/4_experiments.tex
We evaluate ExplicitIR on the three-task and five-task benchmarks, and ask whether the explicit fields match the accuracy of implicit adaptation while tracking where the degradation sits. The efficiency analysis is deferred to Appendix~C.

\begin{table*}[t]
\centering
\caption{Comparison with state-of-the-art on the three-task setting: dehazing on SOTS, deraining on Rain100L, and denoising on CBSD68 (\(\sigma{=}15,25,50\)); PSNR (dB) / SSIM~\cite{wang2004ssim} on full RGB images. The first column separates blind methods, which restore from the degraded image alone, and externally-guided ones, which receive an additional signal at test time.}
\label{tab:main3}
\small
\setlength{\tabcolsep}{4pt}
\begin{tabular}{clcccccc}
\toprule
Category & Method & Dehazing & Deraining & \multicolumn{3}{c}{Denoising (CBSD68)} & Average \\
 & & SOTS & Rain100L & \(\sigma{=}15\) & \(\sigma{=}25\) & \(\sigma{=}50\) & \\
\midrule
\multirow{7}{*}{Blind} & DL~\cite{fan2021dl} & 26.92 / 0.931 & 32.62 / 0.931 & 33.05 / 0.914 & 30.41 / 0.861 & 26.90 / 0.740 & 29.98 / 0.875 \\
 & FD-GAN~\cite{dong2020fdgan} & 24.71 / 0.924 & 29.89 / 0.933 & 30.25 / 0.910 & 28.81 / 0.868 & 26.43 / 0.776 & 28.02 / 0.883 \\
 & MPRNet~\cite{zamir2021multi} & 25.28 / 0.954 & 33.57 / 0.954 & 33.54 / 0.927 & 30.89 / 0.880 & 27.56 / 0.779 & 30.17 / 0.899 \\
 & Restormer~\cite{zamir2022restormer} & 30.43 / 0.975 & 36.55 / 0.975 & 33.84 / 0.931 & 31.18 / 0.885 & 27.90 / 0.790 & 31.98 / 0.911 \\
 & AirNet~\cite{li2022airnet} & 27.94 / 0.962 & 34.90 / 0.967 & 33.92 / 0.933 & 31.26 / 0.888 & 28.00 / 0.797 & 31.20 / 0.910 \\
 & PromptIR~\cite{potlapalli2023promptir} & 30.58 / 0.974 & 36.37 / 0.972 & 33.98 / 0.933 & 31.31 / 0.888 & 28.06 / 0.799 & 32.06 / 0.913 \\
 & AdaIR~\cite{cui2025adair} & 31.06 / 0.980 & 38.64 / 0.983 & 34.12 / 0.935 & 31.45 / 0.892 & 28.19 / 0.802 & 32.69 / 0.918 \\
\midrule
\multirow{4}{*}{Externally-guided} & InstructIR~\cite{conde2024instructir} & 30.22 / 0.959 & 37.98 / 0.978 & 34.15 / 0.933 & 31.52 / 0.890 & 28.30 / 0.804 & 32.43 / 0.913 \\
 & DFPIR~\cite{tian2025dfpir} & 31.87 / 0.980 & 38.65 / 0.982 & 34.14 / 0.935 & 31.47 / 0.893 & 28.25 / 0.806 & 32.88 / 0.919 \\
 & DRNet~\cite{li2026drnet} & 31.15 / 0.979 & 38.28 / 0.983 & 34.20 / 0.937 & 31.55 / 0.894 & 28.27 / 0.807 & 32.69 / 0.920 \\
 & R2R~\cite{wang2026r2r} & 31.40 / 0.977 & 37.46 / 0.980 & 34.10 / 0.936 & 31.45 / 0.895 & 28.22 / 0.806 & 32.53 / 0.918 \\
\midrule
Blind & \textbf{ExplicitIR} & \textbf{31.17 / 0.981} & \textbf{38.88 / 0.985} & \textbf{34.15 / 0.932} & \textbf{31.49 / 0.888} & \textbf{28.21 / 0.799} & \textbf{32.78 / 0.917} \\
\bottomrule
\end{tabular}
\end{table*}

\begin{table*}[t]
\centering
\caption{Comparison with state-of-the-art on the five-task setting: dehazing on SOTS, deraining on Rain100L, denoising on CBSD68 (\(\sigma{=}25\)), deblurring on GoPro, and low-light enhancement on LOL-v1. A star ($^{*}$) marks a general-purpose restorer trained separately for each task; the remaining methods are all-in-one models.}
\label{tab:main5}
\small
\setlength{\tabcolsep}{4pt}
\begin{tabular}{clcccccc}
\toprule
Category & Method & Dehazing & Deraining & Denoising & Deblurring & Low-light & Average \\
 & & SOTS & Rain100L & \(\sigma{=}25\) & GoPro & LOL & \\
\midrule
 & DGUNet$^{*}$~\cite{mou2022dgunet} & 24.78 / 0.940 & 36.62 / 0.971 & 31.10 / 0.883 & 27.25 / 0.837 & 21.87 / 0.823 & 28.32 / 0.891 \\
 & SwinIR$^{*}$~\cite{liang2021swinir} & 21.50 / 0.891 & 30.78 / 0.923 & 30.59 / 0.868 & 24.52 / 0.773 & 17.81 / 0.723 & 25.04 / 0.835 \\
 & Restormer$^{*}$~\cite{zamir2022restormer} & 24.09 / 0.927 & 34.81 / 0.962 & 31.49 / 0.884 & 27.22 / 0.829 & 20.41 / 0.806 & 27.60 / 0.881 \\
 & NAFNet$^{*}$~\cite{chen2022nafnet} & 25.23 / 0.939 & 35.56 / 0.967 & 31.02 / 0.883 & 26.53 / 0.808 & 20.49 / 0.809 & 27.76 / 0.881 \\
Blind & DL~\cite{fan2021dl} & 20.54 / 0.826 & 21.96 / 0.762 & 23.09 / 0.745 & 19.86 / 0.672 & 19.83 / 0.712 & 21.05 / 0.743 \\
 & Transweather~\cite{valanarasu2022transweather} & 21.32 / 0.885 & 29.43 / 0.905 & 29.00 / 0.841 & 25.12 / 0.757 & 21.21 / 0.792 & 25.22 / 0.836 \\
 & TAPE~\cite{liu2022tape} & 22.16 / 0.861 & 29.67 / 0.904 & 30.18 / 0.855 & 24.47 / 0.763 & 18.97 / 0.621 & 25.09 / 0.801 \\
 & AirNet~\cite{li2022airnet} & 21.04 / 0.884 & 32.98 / 0.951 & 30.91 / 0.882 & 24.35 / 0.781 & 18.18 / 0.735 & 25.49 / 0.846 \\
 & IDR~\cite{zhang2023idr} & 25.24 / 0.943 & 35.63 / 0.965 & 31.60 / 0.887 & 27.87 / 0.846 & 21.34 / 0.826 & 28.34 / 0.893 \\
 & AdaIR~\cite{cui2025adair} & 30.53 / 0.978 & 38.02 / 0.981 & 31.35 / 0.889 & 28.12 / 0.858 & 23.00 / 0.845 & 30.20 / 0.910 \\
\midrule
\multirow{4}{*}{Externally-guided} & InstructIR~\cite{conde2024instructir} & 27.10 / 0.956 & 36.84 / 0.973 & 31.40 / 0.887 & 29.40 / 0.886 & 23.00 / 0.836 & 29.55 / 0.907 \\
 & DFPIR~\cite{tian2025dfpir} & 31.64 / 0.979 & 37.62 / 0.978 & 31.29 / 0.889 & 28.82 / 0.873 & 23.82 / 0.843 & 30.64 / 0.913 \\
 & DRNet~\cite{li2026drnet} & 31.28 / 0.980 & 38.13 / 0.982 & 31.54 / 0.894 & 29.01 / 0.870 & 22.30 / 0.846 & 30.45 / 0.914 \\
 & R2R~\cite{wang2026r2r} & 30.64 / 0.974 & 36.61 / 0.975 & 31.35 / 0.891 & 30.93 / 0.911 & 22.88 / 0.856 & 30.48 / 0.921 \\
\midrule
Blind & \textbf{ExplicitIR} & \textbf{30.44 / 0.978} & \textbf{37.47 / 0.980} & \textbf{31.45 / 0.885} & \textbf{30.08 / 0.897} & \textbf{23.41 / 0.850} & \textbf{30.57 / 0.918} \\
\bottomrule
\end{tabular}
\end{table*}

\begin{figure*}[t]
\centering
\includegraphics[width=0.85\textwidth]{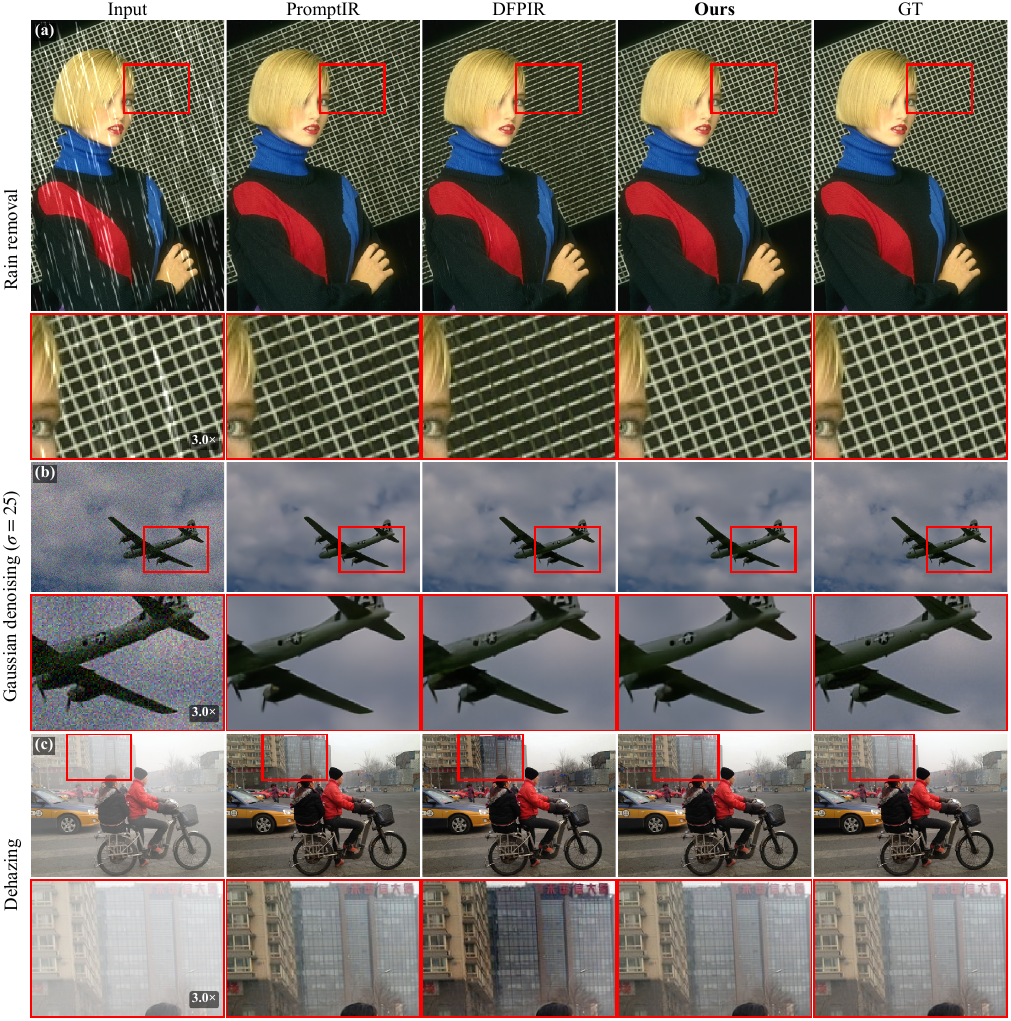}
\caption{Qualitative comparison on the three-task setting for dehazing, deraining, and denoising. ExplicitIR removes the haze, rain streaks, and noise while restoring the underlying colors and structures, and avoids the artifacts of the compared methods. Zoom in for better view.}
\label{fig:qual}
\end{figure*}

\subsection{Experimental Settings}
\label{sec:setup}

\noindent\textbf{Datasets.}
For the three-task setting, we train on the union of BSD400~\cite{arbelaez2011contour} and WED~\cite{ma2017wed} (denoising, with Gaussian noise \(\sigma\in\{15,25,50\}\)), RainTrainL~\cite{yang2017rain} (deraining), and the outdoor OTS subset of RESIDE~\cite{reside2019} (dehazing), and test on CBSD68~\cite{martin2001bsds}, Rain100L~\cite{yang2017rain}, and SOTS-outdoor~\cite{reside2019}, respectively.
The five-task setting adds GoPro~\cite{nah2017gopro} (deblurring) and LOL-v1~\cite{wei2018lol} to both training and testing~\cite{tian2025dfpir,zamfir2025moceir,he2026dacgir}.

\noindent\textbf{Implementation details.}
Training minimizes Eq.~\eqref{eq:total} with AdamW~\cite{loshchilov2019adamw} on the Restormer~\cite{zamir2022restormer} backbone, implemented in PyTorch on a single NVIDIA H100 (80GB).
Crop size, batch size, the number of training epochs, and the learning-rate schedule are listed in Appendix~A.

\noindent\textbf{Baselines.}
The comparison set spans task-specific, general, and all-in-one restorers; the full list and their numbers are given in Tables~\ref{tab:main3} and~\ref{tab:main5}.
Baseline numbers are cited from DFPIR~\cite{tian2025dfpir}, DRNet~\cite{li2026drnet}, and AdaIR~\cite{cui2025adair}, which report the same data splits.
All methods are evaluated as specified in the table captions; ExplicitIR is blind and receives no degradation type or level at test time.

\subsection{Results on Three Tasks}
\label{sec:main3}
Table~\ref{tab:main3} compares ExplicitIR with these baselines under the three-task protocol~\cite{potlapalli2023promptir,tian2025dfpir,zamfir2025moceir,he2026dacgir}.

ExplicitIR reaches 32.78 dB on average, outperforming the strongest blind method and nearly matching the strongest externally-guided method, which receives the degradation type at test time.
The gain is concentrated on deraining.
Rain streaks are sparse and localized, so Budget can confine the strong update to the streak pixels and leave the flat background untouched; a method that conditions the whole image on one rain representation cannot make that distinction, which is where the margin comes from.
The relative gain is smaller for dehazing: since haze is a global intensity shift and the ideal correction is close to uniform, the Budget field has less spatial structure to exploit. Consequently, ExplicitIR trails the best externally-guided baseline here by a small margin.
Denoising falls in between: since noise is spread everywhere, Budget applies broad updates across the image, while Style favors narrow windows to remove fine details. Consequently, ExplicitIR remains highly competitive with DFPIR at both \(\sigma{=}25\) and \(\sigma{=}50\).

\subsection{Results on Five Tasks}
\label{sec:main5}
Table~\ref{tab:main5} extends the evaluation to five tasks; denoising uses \(\sigma{=}25\), following DFPIR~\cite{tian2025dfpir}.
Deblurring is the informative addition: its useful context is the widest of the five tasks, so a Style field that trades window width should show its clearest effect here.

ExplicitIR reaches 30.57 dB on average, demonstrating strong generalization and maintaining a clear lead over blind methods.
The gain is concentrated on deblurring, where ExplicitIR significantly outperforms both AdaIR and InstructIR.
A blurred pixel needs a wide, edge-preserving update, and a Style field that can select the wide window per location supplies exactly that, whereas methods with a fixed window width apply the same context everywhere.
The remaining columns split between the two families: IDR leads on denoising, where wide context helps less and Style has little to select; DFPIR leads on dehazing and low-light enhancement, two tasks whose correction is a near-global intensity change, the regime where per-location strength has the least to add; DRNet leads on deraining, which is the strongest column in the three-task setting but not here, consistent with the five-task model reallocating capacity toward the two added tasks rather than with a change in the fields themselves.
We report all entries as measured and note that DFPIR receives a degradation-type description at test time, while IDR, AdaIR, and ExplicitIR are blind.

\subsection{Visual Results and Policy Fields}
\label{sec:visual}

\begin{figure*}[t]
\centering
\includegraphics[width=0.96\textwidth]{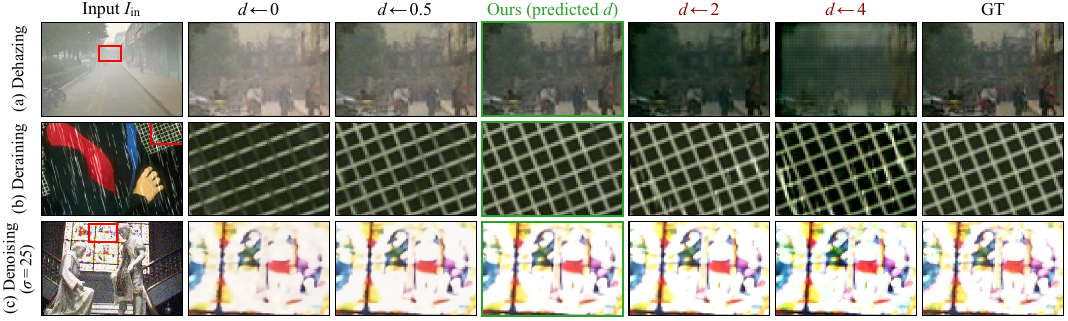}
\caption{Editing the Budget field at inference. The columns show the degraded input, re-decodes with the Budget in the marked region clamped to 0 and 0.5, the model prediction, and forced to 2 and 4, and the clean reference. The restoration improves as the Budget rises toward the predicted value, but degrades when forced beyond the valid range \((0,1)\), demonstrating that the field directly controls the correction strength.}
\label{fig:intervention}
\end{figure*}

\begin{figure*}[t]
\centering
\includegraphics[width=0.96\textwidth]{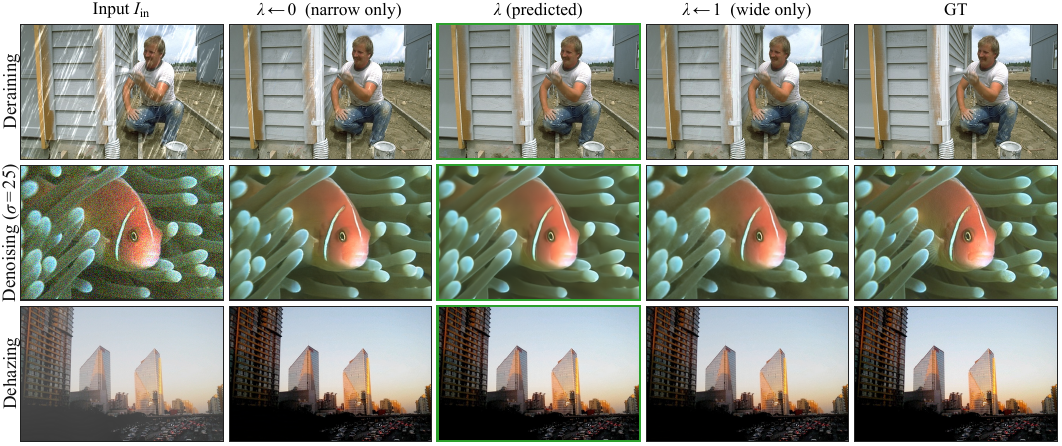}
\caption{Editing the Style field globally at inference. The columns show the degraded input, re-decodes with the Style fixed to 0 (narrow window only) and 1 (wide window only), the model prediction, and the clean reference. Clamping to either extreme degrades restoration in a task-characteristic way: narrow-only leaves streaks and grain, while wide-only over-smooths the texture.}
\label{fig:intervention_lambda}
\end{figure*}

Fig.~\ref{fig:qual} presents qualitative comparisons on the three-task setting.
Compared to the baselines, ExplicitIR more faithfully restores structures in heavily corrupted regions while avoiding over-smoothing or artifacts in clean areas. 
For dehazing, it avoids the over-smoothing artifacts commonly introduced by globally uniform restoration operations, yielding sharper edges in the distant regions without altering the near-field details. 
This spatial adaptivity is visually evident in the varying update strength across a single image, demonstrating the advantage of the learned per-pixel fields over globally uniform restoration.

To further establish a causal link between the learned fields and the restoration outcome, we perform inference-time interventions: we explicitly edit a field and re-decode the image, forcing the restoration to follow the edited value rather than the network's own prediction.

\noindent\textbf{Budget interventions.}
Figure~\ref{fig:intervention} manipulates the Budget field inside the marked region. The restoration quality peaks near the model's predicted Budget and drops when the value is clamped to 0 or forced to 2 and 4. Notably, since \(d = \sigma(\cdot) \in (0,1)\), forcing \(d\) to 2 or 4 breaks the RMS bound in Eq.~\eqref{eq:bound}, resulting in severe over-correction into over-smoothed textures. This not only confirms that Budget governs the update magnitude but also validates the necessity of our bounded injection design.

\noindent\textbf{Style interventions.}
Figure~\ref{fig:intervention_lambda} overrides the Style field globally, forcing every location to exclusively use the narrow (\(\lambda\equiv 0\)) or the wide (\(\lambda\equiv 1\)) window. Both extremes degrade the restoration in a task-characteristic way, leaving the predicted value as the optimal operating point. Forcing \(\lambda\equiv 0\) or \(\lambda\equiv 1\) both lead to a significant drop in PSNR, confirming that the network benefits from dynamically selecting the context window per location rather than applying a single global scale.

Additional field visualizations and Style distributions are provided in Appendix~B.

\subsection{Ablation Study and Policy Diagnostics}
\label{sec:ablation}
We conduct ablations on the three-task setting; detailed variant definitions and full results are provided in Appendix~D.
Two findings validate our design. First, the policy path is highly synergetic: removing both modules drops the average to 31.98 dB, whereas keeping them individually recovers only 0.47 dB combined, falling well short of the full 0.80 dB gain. Second, compared to an implicit end-to-end gating baseline, ExplicitIR achieves slightly better accuracy while maintaining a significantly higher Budget mask-gap (the difference in allocated Budget between corrupted and clean regions), demonstrating that our explicit readable form provides both competitive restoration and measurable spatial allocation.

%% file: sections/5_conclusion.tex
In this paper, we propose ExplicitIR, which represents the adaptation policy of all-in-one restoration as two explicit per-pixel fields, Budget and Style.
A Spatial Policy Estimator predicts both fields from the degraded image alone, and Policy-Adaptive Decoder Blocks apply them under bounded gating.
Extensive experiments demonstrate that ExplicitIR achieves strong performance, approaching the strongest externally-guided methods despite requiring no external guidance at inference.
Future work will extend this framework to real-world degradations and diverse backbones.

%% file: appendix.tex

\input{sections/A_appendix}

\begin{figure*}[t]
\centering
\includegraphics[width=0.95\textwidth]{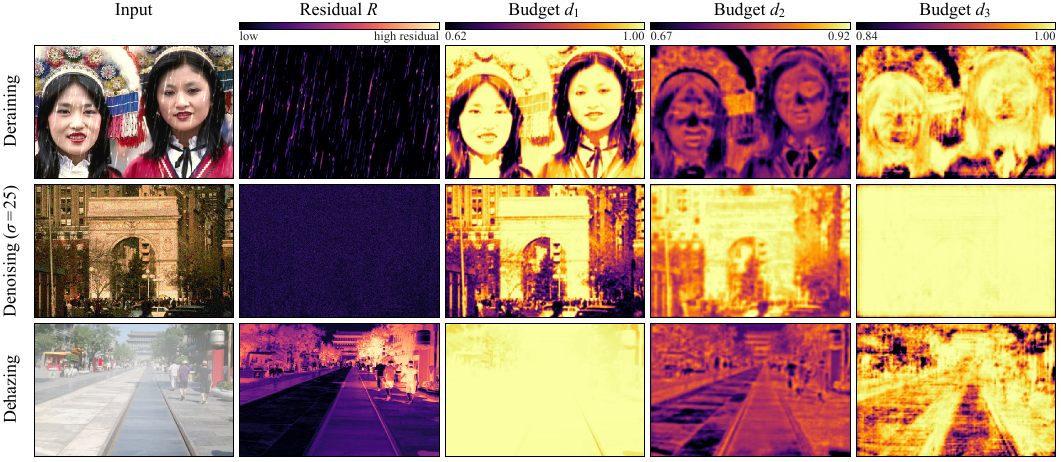}
\caption{Budget fields across decoder stages. Each row shows one degradation; the columns give the degraded input, the residual against the clean reference, and the Budget fields at stages 1 to 3, with each stage shown on its own scale and the numeric range indicated above each column. The allocation coarsens with depth: stage 1 follows fine image structure, while the deeper stages move to region-level updates.}
\label{fig:abl_d}
\end{figure*}

\begin{figure*}[t]
\centering
\includegraphics[width=0.95\textwidth]{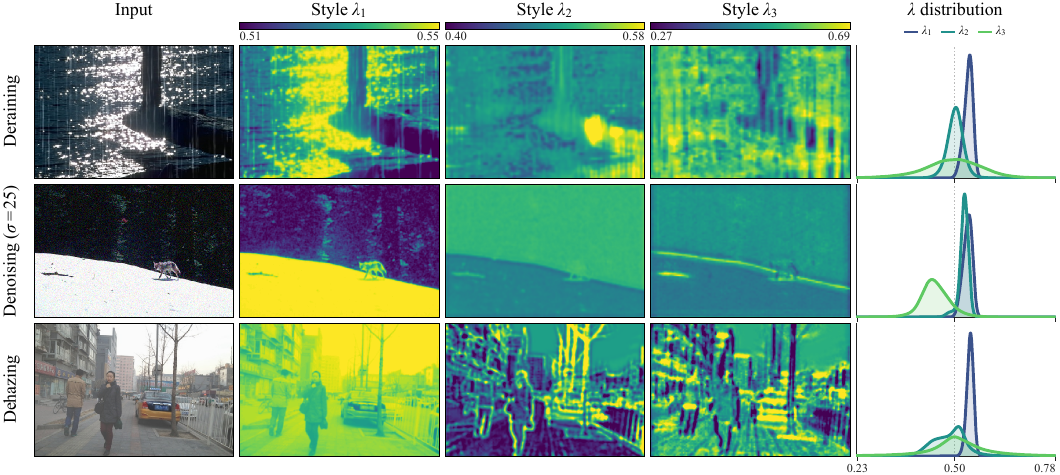}
\caption{Style fields across decoder stages, with the pooled Style distribution over each test set (rightmost column). Each row shows one degradation. The predicted Style stays close to the balanced point at stage 1 and follows image structure at the deeper stages.}
\label{fig:stylediag}
\end{figure*}

\section{Additional Field Visualizations}
\label{app:fields}
This section inspects the two predicted fields on the three-task model.
Figures~\ref{fig:abl_d} and~\ref{fig:stylediag} show the spatial structure of Budget and Style across decoder stages.

\noindent\textbf{Budget fields.}
Figure~\ref{fig:abl_d} visualizes the Budget fields across the decoder stages: each row is one degradation, and the columns give the degraded input, the residual against the clean reference, and the Budget fields at stages 1 to 3, with each stage shown on its own scale, where brighter means a stronger update.
Stage 1 follows fine image structure, and the allocation coarsens with depth into region-level updates, tracking the growth of the receptive field.
The three fields are not redundant: setting the stage-1 Budget to zero changes the output by a larger margin than setting the stage-2 or stage-3 Budget to zero on all three test sets, so the first field carries most of the correction strength and the deeper fields refine where it concentrates.

\noindent\textbf{Style fields.}
Figure~\ref{fig:stylediag} visualizes the Style fields analogously: the first three columns give the Style fields at the three decoder stages, and the rightmost column pools the Style values of the whole test set.
At stage 1 the predicted Style stays close to the balanced point, and the deeper stages follow image structure.
The pooled distributions occupy different ranges across the three tasks, indicating that the two fields are separately controlled rather than redundant.

\section{Efficiency Analysis}
\label{app:efficiency}
\begin{figure}[t]
\centering
\includegraphics[width=\columnwidth]{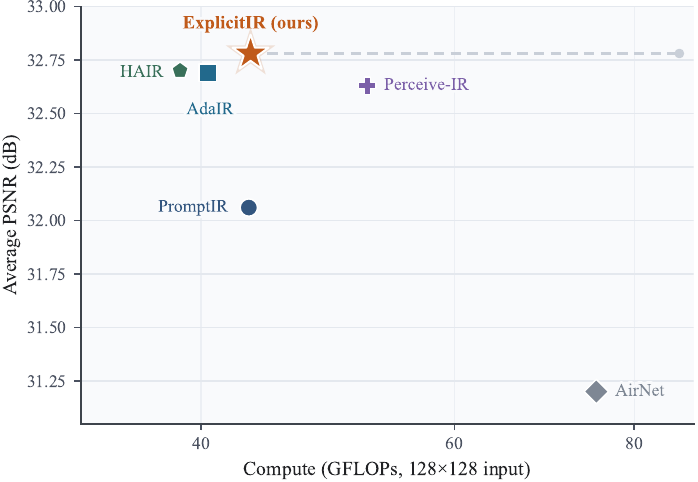}
\caption{Accuracy versus compute on the three-task setting: average PSNR against the number of operations measured on the official implementations at a \(128{\times}128\) input. ExplicitIR is marked with a star, and the dashed line tracks its average.}
\label{fig:efficiency}
\end{figure}

Fig.~\ref{fig:efficiency} plots average PSNR against compute on the three-task setting.
ExplicitIR reaches the highest average of this band at moderate compute: PromptIR~\cite{potlapalli2023promptir} spends nearly the same compute (43.2 versus 43.3 GFLOPs) at a 0.72 dB lower average, AirNet~\cite{li2022airnet} spends 75.3 GFLOPs, about \(1.7\times\) the compute of ExplicitIR, and remains 1.58 dB below it, and the lighter AdaIR~\cite{cui2025adair} lies within 0.09 dB below ExplicitIR at 36.6 GFLOPs.

\section{Additional Ablations}
\label{app:ablation}
To quantify whether the Budget field successfully localizes the corruption, we introduce the \textbf{Mask-gap} metric. For each test image pair, we synthesize a half-corrupted input by concatenating the left half of the degraded image with the right half of its corresponding clean image. The Mask-gap is computed as the mean Budget value on the corrupted half minus that on the clean half. A larger Mask-gap indicates a more precise spatial allocation of update strength, whereas a zero gap implies a spatially uniform policy.

\begin{table}[t]
\centering
\caption{Ablations on the three-task setting, grouped by the claim each variant tests. \textit{Module}: whether the gain comes from backbone capacity, field prediction, or field execution; Backbone only removes SPE and PADB, SPE only predicts without execution, and PADB only executes without prediction. \textit{Separation}: whether the two fields contribute independently and whether the readable form carries the gain; Free gating keeps per-pixel capacity but removes the readable form, and Budget-only / Style-only fix the other field. \textit{Variation}: whether spatial variation matters beyond a global amplitude change, and whether the dual-window design is necessary. Every variant is trained with the reconstruction loss of Eq.~(6).}
\label{tab:ablation}
\small
\setlength{\tabcolsep}{3pt}
\begin{tabular}{@{}lcc@{}}
\toprule
Variant & Avg PSNR & Mask-gap \\
\midrule
\multicolumn{3}{@{}l}{\textit{Module effectiveness}} \\
Full ExplicitIR & 32.78 & 0.14 \\
Backbone only & 31.98 & n/a \\
SPE only & 32.08 & 0.03 \\
PADB only & 32.35 & 0.00 \\
\midrule
\multicolumn{3}{@{}l}{\textit{Field separation}} \\
Free gating & 32.70 & 0.08 \\
Budget-only & 32.62 & 0.13 \\
Style-only & 32.42 & 0.00 \\
\midrule
\multicolumn{3}{@{}l}{\textit{Spatial variation and window design}} \\
Uniform policy & 32.38 & 0 \\
Amplitude-matched \(d\!\equiv\!\bar{d}\) & 32.50 & 0.00 \\
Single MDTA window & 32.60 & 0.10 \\
\bottomrule
\end{tabular}
\end{table}

Table~\ref{tab:ablation} separates the contributions of the two modules, the two fields, and the spatial variation.
Removing SPE and PADB together, \textit{Backbone only} reaches 31.98 dB, so the policy path accounts for 0.80 dB of the full model.
Adding only SPE (\textit{SPE only}, 32.08 dB) recovers 0.10 dB, and adding only PADB (\textit{PADB only}, 32.35 dB) recovers 0.37 dB; the combined gain of 0.80 dB exceeds the sum of the individual gains (0.47 dB), indicating strong synergy between the predicted fields and the execution mechanism.
The separation rows show that the explicit readable form carries a gain beyond per-pixel capacity: \textit{Free gating} keeps end-to-end gates and reaches 32.70 dB, only 0.08 dB below the full model, but its mask-gap drops from 0.14 to 0.08. The explicit fields achieve slightly better accuracy while keeping the spatial allocation measurable.
Within the two fields, \textit{Budget-only} reaches 32.62 dB with a mask-gap of 0.13, close to the full model, whereas \textit{Style-only} reaches 32.42 dB with a mask-gap of 0.00 (since \(d\) is fixed to 1); Budget therefore carries most of the spatial correction, and Style contributes 0.16 dB on top of it.
The variation rows separate spatial variation from a global amplitude change: \textit{Uniform policy} drops to 32.38 dB with a mask-gap of zero by construction, and the amplitude-matched control \(d\!\equiv\!\bar{d}\) recovers 0.12 dB to 32.50 dB with a zero mask-gap by definition, but remains 0.28 dB below the full model, so the gain cannot be explained by global amplitude alone.
Replacing the two MDTA windows with a single window lowers the average to 32.60 dB, which supports the dual-window design.

%% file: sections/A_appendix.tex
\section{Model and Training Configuration}
\label{app:impl}
Table~\ref{tab:hparams} lists the model architecture and training configuration; a single setting serves the three-task and five-task benchmarks.
The per-stage constant \(\kappa^{(l)}\) is fixed before training as \(\kappa^{(l)}=k_l\,r_l\,\rho_l\).
Here \(k_l=1\) is a constant multiplier, \(r_l\) is the ratio of a backbone block's residual output magnitude to its input magnitude at stage \(l\) (averaged over the stage's blocks), and \(\rho_l\) is the per-location cross-channel RMS of the decoder feature \(G_l\) at initialization.
Both are measured on \(128\times128\) degraded inputs, with \(r_l\) taken from the frozen Restormer checkpoint and \(\rho_l\) at initialization.

\begin{table}[t]
\centering
\caption{Model architecture and training configuration.}
\label{tab:hparams}
\small
\setlength{\tabcolsep}{4pt}
\begin{tabular}{@{}p{0.54\columnwidth}p{0.38\columnwidth}@{}}
\toprule
Item & Value \\
\midrule
\multicolumn{2}{@{}l}{\textit{Architecture}} \\
Backbone & Restormer~\cite{zamir2022restormer} \\
Backbone blocks per stage & \([4,6,6,8]\) \\
Attention heads per stage & \([1,2,4,8]\) \\
Encoder embedding dim & \([48, 96, 192, 384]\) \\
Decoder stages with PADB & 3 \\
Narrow MDTA window & 8 \\
Wide MDTA window & 32 \\
SPE hidden channels & 64 \\
RMS stabilizer \(\epsilon\) & \(10^{-6}\) \\
\midrule
\multicolumn{2}{@{}l}{\textit{Training}} \\
Optimizer & AdamW \\
\(\beta_1, \beta_2\) & \(0.9\), \(0.999\) \\
Learning rate & \(2\times10^{-4}\) \\
Learning-rate schedule & Cosine to zero, 10\% linear warmup \\
Weight decay & \(10^{-2}\) \\
Batch size & 32 effective (16 physical, 2 accumulation steps) \\
Crop size & \(128\times128\) \\
Training epochs & 150 \\
Loss & \(L_1\), RGB domain \\
\bottomrule
\end{tabular}
\end{table}

\subsection{Windowed MDTA}
\label{app:mdta}
Both forms are the MDTA of Restormer~\cite{zamir2022restormer} applied within non-overlapping tiles instead of the whole feature map.
Let \(\mathcal{W}_{w}(H,W)\) partition the spatial grid into non-overlapping tiles of side \(w\).
On a tile \(\Omega\in\mathcal{W}_{w}(H,W)\), the operator uses only the \(|\Omega|\) positions of that tile,
\begin{equation}
\label{eq:mdta-window}
A_{\Omega}=\mathrm{softmax}\!\Big(\frac{Q_{\Omega}K_{\Omega}^{\top}}{\beta}\Big),
\qquad
\mathrm{out}_{\Omega}=A_{\Omega}V_{\Omega},
\end{equation}
and the narrow and wide forms take \(w_{\mathrm{nar}}=8\) and \(w_{\mathrm{wid}}=32\).

Because MDTA attends across channels, restricting the positions changes the support of the spatial statistics but does not alter the theoretical computational complexity, which remains \(O(N(C/h)^2)\) for the entire partition. Consequently, evaluating two windowed branches per block approximately doubles the attention computation compared to a single full-map operator. However, since the channel dimension \(C/h\) is relatively small, this attention module contributes minimally to the overall FLOPs. The primary advantage of windowing lies in reducing the peak memory footprint during the \(Q K^\top\) matrix multiplication and injecting a local inductive bias, while the dual branches maintain a negligible overall overhead (as reflected in the efficiency analysis of Appendix C).

\subsection{RMS normalization}
\label{app:rms}
Eq.~(3) normalizes with \(\mathrm{norm}(X)=X/\mathrm{RMS}(X)\), where \(\mathrm{RMS}(X)_{i,j}\) is the root mean square of the channel values at location \((i,j)\), computed with a small stabilizer \(\epsilon=10^{-6}\).
The statistic is taken over channels at a single location, as in the layer normalization of the backbone, but no mean is subtracted.